\documentclass{article}

\usepackage{indentfirst}
\usepackage{graphicx}
\usepackage{amsmath}
\usepackage{subcaption}
\usepackage{amsfonts}
\usepackage{amssymb}
\usepackage{algorithm}
\usepackage{algorithmic}
\usepackage{booktabs}
\usepackage{array}
\usepackage{adjustbox}
\usepackage[symbol]{footmisc}
\usepackage{multirow}
\usepackage[final]{corl_2020} 

\def\ie{\emph{i.e.}}

\def\etc{\emph{etc}}
\newcommand{\etal}{\textit{et al.}}

\title{Cylinder3D: An Effective 3D Framework for Driving-scene LiDAR Semantic Segmentation}

\author{
  Hui Zhou$^{*\S}$~~~~~ Xinge Zhu$^{*\dag}$~~~~~ Xiao Song$^{\S}$~~~~~ Yuexin Ma$^{\ddagger}$ 
  \\ \textbf{Zhe Wang}$^{\S}$~~~~~ \textbf{Hongsheng Li}$^{\dag}$~~~~~ \textbf{Dahua Lin}$^{\dag}$ \\
  $^{\dag}$The Chinese University of Hong Kong~~~~$^{\ddagger}$ShanghaiTech University\\$^{\S}$SenseTime Research}

\begin{document}
\maketitle


\begin{abstract}
    
State-of-the-art methods for large-scale driving-scene LiDAR semantic segmentation often {project and process the point clouds in the 2D space}. {The projection methods} includes spherical projection, bird-eye view projection, \etc. Although this process makes the point cloud suitable for the 2D CNN-based {networks}, it  inevitably alters {and abandons} the 3D topology and geometric relations. 
A straightforward {solution} to tackle the issue of 3D-to-2D projection is to keep the 3D representation and process the points in the 3D space. 
In this work, we first perform an in-depth analysis for different representations and backbones in 2D and 3D spaces, and reveal the effectiveness of 3D representations and networks on LiDAR segmentation. Then, we develop a 3D cylinder partition and a 3D cylinder convolution based framework, termed as Cylinder3D, which exploits the 3D topology relations and structures of driving-scene point clouds. 
Moreover, a dimension-decomposition based context modeling module is introduced to explore the high-rank context information in point clouds in a progressive manner.
We evaluate the proposed model on a large-scale driving-scene dataset, \ie~SematicKITTI. Our method achieves state-of-the-art performance and outperforms existing methods by 6\% in terms of mIoU.
    
\end{abstract}

\keywords{Cylinder Partition, Asymmetric Residual Block, Context Modeling
\footnote[1]{equal contribution. Source code: \url{https://github.com/xinge008/Cylinder3D}}} 


\section{Introduction}
	 3D {LiDAR} sensor has become an {indispensable device} in {modern} autonomous driving {vehicles}. It {captures} more precise and further-away {distance measurements of the surrounding environments} than conventional visual cameras. {The measurements of the sensor naturally form 3D point clouds that can be used to understand the overall scenes for autonomous driving planning and execution.}
	 
	 
	 {Semantic segmentation of 3D point clouds is crucial for {driving-}scene understanding. It aims to identify the pre-defined categories of each 3D point that belongs to, such as {car, truck, pedestrian,} \etc, which provides point-wise perception information of the overall 3D scene.}
	
    
    Most existing point cloud-based segmentation algorithms {focus more on} indoor scenes, {where the point clouds are generally dense and have mostly uniform {densities}.} In contrast, only
    very few {methods} {work} on {segmentation of LiDAR point clouds in outdoor or autonomous driving scenes, where the LiDAR points have varying densities according to their distances to the sensor and pose {great} challenges to the algorithms.} 
   
    Recent methods usually pay much attention on point feature representations \citep{wu2018squeezeseg, zhang2020polarnet, graham20183d}. Point feature representation for LiDAR point clouds has three major {categories}: range image \citep{milioto2019rangenet++,wu2018squeezeseg}, bird view
   image \citep{zhang2020polarnet} and voxel partition \citep{graham20183d, cciccek20163d}. The range image {is obtained via} spherical projection of the irregularly distributed 3D point clouds to the 2D dense grids. The bird-view image squeezes point height information and shares a global height feature for each location on the bird view map. However, most of these approaches {might lose certain accurate geometric information during the {3D-to-2D} projection.}

    \begin{figure*}
    \centering
    \includegraphics[width=1.0\linewidth]{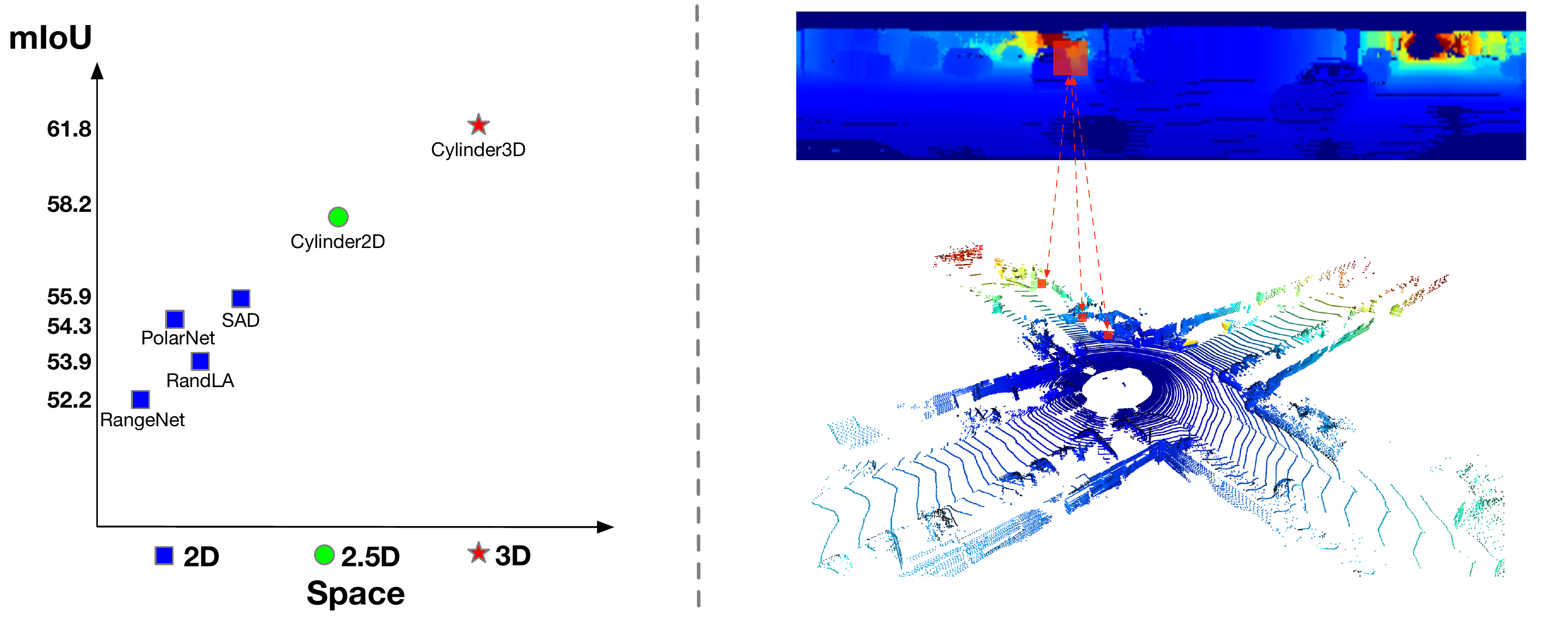}
    \caption{{(Left)} The detailed road map for {network architecture search on SemanticKITTI}, from 2D, 2.5D to 3D (Note that 2.5D means 3D grid representation and 2D backbone). {(Right)} The limitation of spherical projection, namely, {abandons certain valuable 3D structures}, where neighboring region in projection reflects significantly different locations in 3D space, which shows that spherical projection cannot maintain the 3D geometry structure.}
    \label{fig:3d_conv}
    \end{figure*}

     In this paper, we reposition the focus of LiDAR segmentation in autonomous driving scenes.
     This paper conducts experiments to show effectiveness of different point feature representations and neural network architectures. Experiments reveal that 3D partition with 3D convolutional neural networks works better than other counterparts. A cylinder partition is proposed to process the driving-scene point clouds due to its varying densities, which balances the distribution of driving-scene point clouds. To match the cuboid objects in driving-scene LiDAR data, we propose the asymmetric residual block as a basic module to form the 3D backbone.
     {In addition to the network search, we also propose a new dimension decomposition block to efficiently exploit the context information via a series of low-rank convolution kernels.}
    
     The contributions of this work {can be summarized as three-fold. (1) We study state-of-the-art network architectures and different point feature representations, which reveal {directly processing point clouds without 3D-to-2D projection is crucial for achieving superior segmentation performance.} (2) We propose a cylinder partition, {a point cloud encoding scheme, which better follows the inherent distribution of the 3D driving-scene point clouds}, and develop a 3D convolution based framework, in which the asymmetric residual block is designed as the basic module and a new dimension decomposition block is proposed to explore the context in a progressive manner. (3) Our proposed LiDAR segmentation algorithm outperforms state-of-the-art algorithms on {driving-scene} semantic segmentation benchmarks with a large margin, \ie,~6\% mIoU gain.}

        
        

\section{Related Work}
   
   
   \textbf{Indoor-scene Point Cloud Segmentation.} Indoor-scene point clouds have some properties, including generally uniform density and small range of the scene. Hence, most indoor-scene segmentation methods \citep{qi2017pointnet, thomas2019kpconv, wu2019pointconv, wang2019dynamic,velivckovic2017graph, lyu2020learning, engelmann20203d, zhang2020fusion} often learn the point features from the raw point directly. PointNet \citep{qi2017pointnet} is a classical convolutional neural network on point sets and proposed a multi-layer perception to extract features from input points. Moreover, PointNet++ \citep{qi2017pointnet++} further proposed multi-scale sampling to aggregate global and local features. Another group of indoor-scene segmentation~\cite{wang2019dynamic,velivckovic2017graph} utilizes the clustering (including KNN) to extract the point features. 
   However, these methods are computationally costly and do not take varying sparsity (the property of outdoor-scene LiDAR) into consideration.

   \textbf{Outdoor-scene Point Cloud Segmentation.}
   Most existing outdoor-scene point cloud segmentation focuses on converting the 3D point cloud to 2D grids to enable the use of 2D Convolutional Neural Networks. SqueezeSeg \citep{wu2018squeezeseg}, Darknet \citep{behley2019semantickitti}, SqueezeSegv2 \citep{wu2019squeezesegv2},
  and RangeNet++ \citep{milioto2019rangenet++} utilize the spherical projection mechanism, which converts the point cloud to 
  a frontal-view (range) image, and adopt the 2D convolution network on the pseudo image for segmentation. PolarNet \citep{zhang2020polarnet} follows the bird-view projection, which projects point cloud data into small grids from the bird view and takes the height as a whole. Instead of partitioning points in a Cartesian coordinate system, they use a polar coordinate system for encoding point clouds. However, this 3D-to-2D projection inevitably compresses the 3D topology and fails to model the geometric information.
  
  \textbf{3D Voxel Partition}
  3D voxel partition is another routine of point cloud encoding~\cite{wang2020reconfigurable,han2020occuseg,zhu2020ssn,tchapmi2017segcloud,graham20183d}. It converts a point cloud into 3D voxels. 3D U-Net \citep{cciccek20163d} proposes voxel partition and 3D U-Net on biomedical data and shows successful application on difficult microscopic datasets. OccuSeg \citep{han2020occuseg}, SSCN \citep{graham20183d} and SEGCloud \citep{tchapmi2017segcloud} follow this line to utilize the voxel partition and apply 3D convolutions for LiDAR segmentation. Our work also follows this routine, utilizing the 3D grid and 3D convolution networks, but with substantial differences. We use the 3D cylinder partition based on the cylinder coordinate system, which meets the varying sparsity of driving-scene LiDAR point cloud and balances point distribution. Specifically, distant region performs much sparse than closer one, and cylinder partition thus utilizes a larger cylinder to cover the distant region accordingly.

\textbf{Network Architectures for Segmentation}. Fully Convolutional Network \citep{long2015fully} is the fundamental work in the deep-learning era. U-Net \citep{ronneberger2015u} built upon FCN and proposed a symmetric architecture to utilize the low-level features. Furthermore, many works explore the dilated convolution for multi-scale context modeling, including DeepLab\citep{chen2017deeplab, chen2018encoder} and PSP~\cite{zhao2017pyramid}.
Due to the great success of U-Net on 2D benchmarks, many studies for LiDAR segmentation adapt the U-Net to the 3D space and propose 3D U-Net \citep{cciccek20163d}. However, they often fail to explore the distribution and property of the driving-scene LiDAR point cloud. In this work, two modules, \ie, Asymmetric Residual Block and Dimension-decomposition based Context Modeling, are designed to match the cuboid objects and model the high-rank context information, respectively.



\section{Methodology}
\label{sec:methods}


\subsection{Study on 3D Point Representations}

Outdoor-scene point clouds have significant differences with indoor-scene point clouds. (1) A {driving-scene} point cloud might cover a very large area, as far as over 100 meters. (2) It generally contains more points ($>$100,000 points) but are much sparser than those of the indoor scenes. Hence, the indoor segmentation methods working on dense and fixed-number points {are} difficult to be adapted to the driving scenes with varying point densities.

Existing outdoor LiDAR segmentation methods mainly focus on transforming the 3D point clouds to 2D representations via projection, including spherical projection and bird-eye view projection, and then adopt 2D convolutions to process the 2D grid representations. However, as shown in Fig.~\ref{fig:3d_conv}(right), the local spatial pattern in 2D grid representation cannot well capture 3D geometric structures. It can be observed that the red rectangle in 2D grid denotes the points distributing in different spatial locations.
Hence, these 3D-to-2D projection methods may fail to encode certain 3D geometric structures and incur inaccurate pattern extraction. The detailed survey is shown in Section~\ref{sec:exp_baseline}. We perform extensive experiments with various partition and networks among 2D, 2.5D and 3D.
From the results, the consistent performance gain indicates the effectiveness of our technical road map, namely, 3D partition and 3D networks.

    \begin{figure*}
    \centering
    \includegraphics[width=1\linewidth]{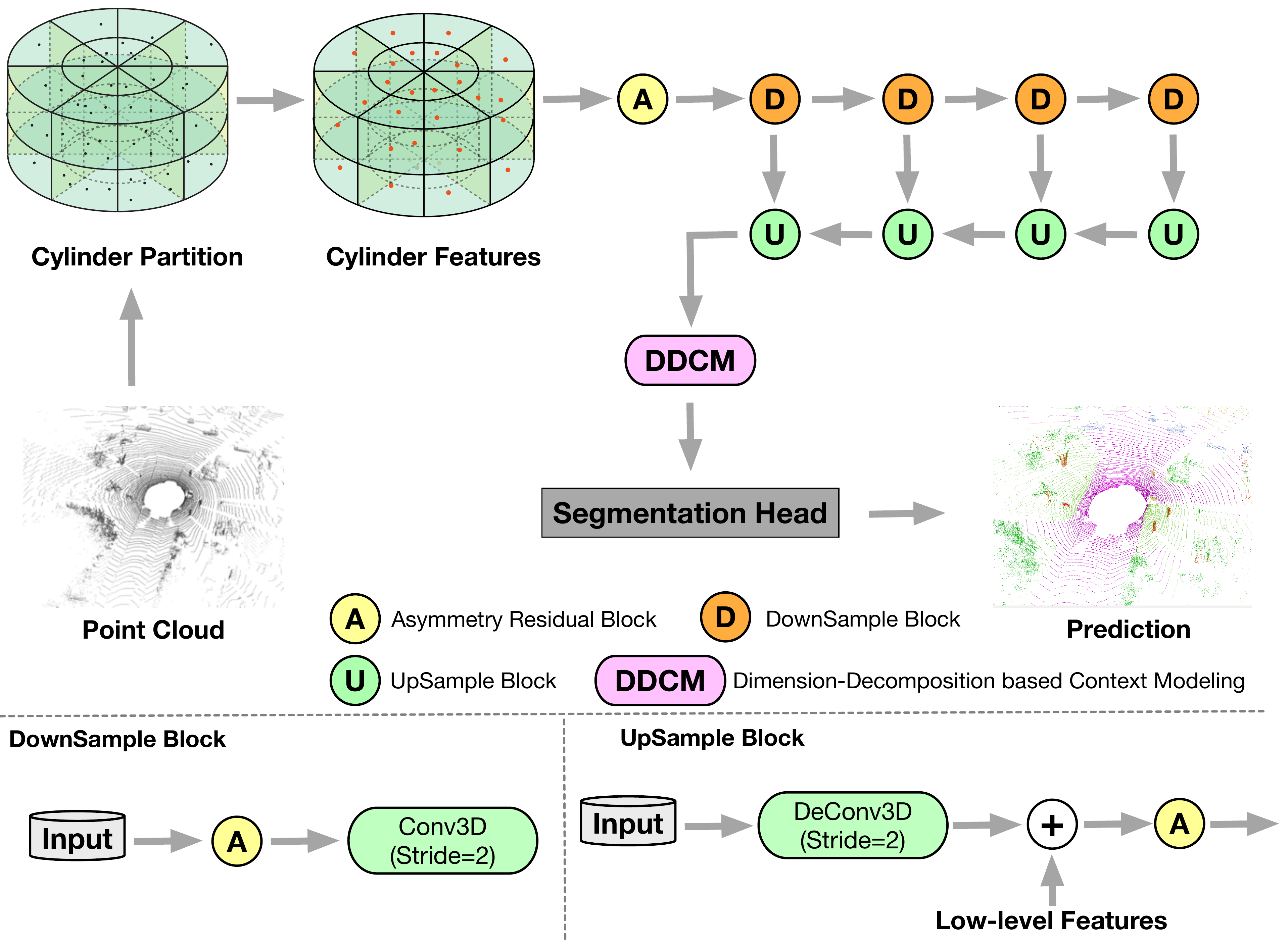}
    \caption{The overall architecture. Top part is the full workflow of the proposed 3D LiDAR segmentation network, Cylinder3D. Bottom parts are the details of the Downsample block and UpSample block.}
    \label{fig:overall}
    \end{figure*}

\subsection{Framework Overview}

The outdoor point clouds are covering a large varieties of urban scenes. Our task is to assign a semantic label to each point in the point cloud. Based on our investigation on the distribution of 2D and 3D point-cloud representations, we discover that the 2D representation obtained from projection would abandon many available 3D structures. To this end, we propose a new outdoor LiDAR segmentation approach based on 3D representation and neural networks. 

 As shown in Fig.~\ref{fig:overall}, the framework consists of two components, including 3D cylinder partition (to obtain the 3D representation) and 3D U-Net (to process the 3D representation). Particularly, we design two modules to suit the properties of outdoor point clouds, \ie, Asymmetrical Residual Block to match these cuboid based objects often appearing in the driving scenes (cars, trucks, motorcycles, \etc), and dimension-decomposition based context modeling module to exploit the high-rank context information in point clouds in a decomposition-aggregation manner. In the following sections, we will introduce these components in detail.

    \begin{figure*}
    \centering
    \includegraphics[width=1\linewidth]{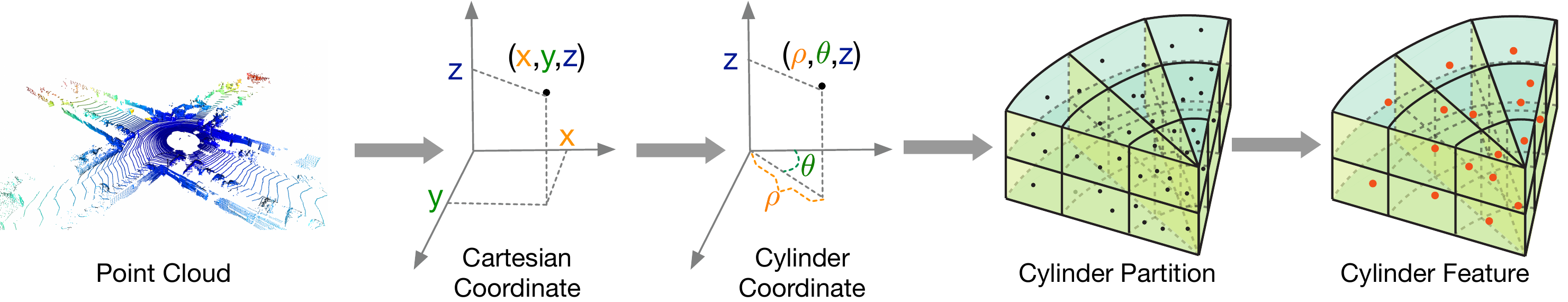}
    \caption{The pipeline of Cylinder Partition. It first transforms points on Cartesian coordinate to Cylinder coordinate. Then a cylinder partition is introduced to perform the voxelization. Finally, cylinder features are produced by a simplified pointnet.}
    \label{fig:cylinder}
    \end{figure*}

\subsection{Cylinder Partition}
As mentioned above, outdoor-scene LiDAR point cloud possesses the property of varying density, where nearby region has much greater density than distant region. We thus use the cylinder coordinate system to replace the Cartesian grid partition. 
It utilizes the increasing grid to cover the further-away region, thus it more evenly distributes the points across different regions and matches the distribution of outdoor points. Moreover, unlike these projection-based methods project the point to the 2D view, we maintain the 3D grid representation to retain the geometric structure. The workflow is shown in Fig.~\ref{fig:cylinder}. We first transform the points on Cartesian coordinate system to the Cylinder coordinate system, where radius $\rho$ and azimuth $\theta$ are calculated. This step transforms the points ($x, y, z$) to points ($\rho, \theta, z$). Then cylinder partition is to split these three dimensions uniformly, note that this split indicates more further-away region, larger voxel. These cylinder grid representation is fed to a MLP-based pointnet to get the cylinder features.
After these steps, we can get the 3D cylinder representation $\mathbb{R}\in C \times H\times W \times L$, where $C$ denotes the feature dimension.

\subsection{Asymmetric Residual Block}


For the autonomous driving scenes, there exist a large amount of cuboid objects, including cars, trucks, buses and motorcycles. Inspired by text detection methods~\cite{wang2019shape}, where asymmetry convolutional kernels are used to match the rectangle target regions, we design the asymmetric residual block to meet the property of such cuboid objects. Moreover, this asymmetric residual block also significantly reduces the computational cost of conventional 3D convolutional kernels.
Specifically, using a convolution with kernel=$3\times1\times3$ followed by a $1\times3\times3$ convolution is equivalent to sliding a two layer network with the same receptive field as in a 3D convolution with kernel= $3\times3\times3$, but it has 33\% cheaper computational cost than a $3\times3\times3$ convolution with same number of output filters. The proposed asymmetrical residual block is the basic component of downsample block and upsample block. For downsample block, it consists of a asymmetrical residual block and a 3D convolution with stride=2 to perform downsample. Upsample block incorporates the low-level features and processes the fused features with a asymmetrical residual block.


    \begin{figure*}
    \centering
    \includegraphics[width=1\linewidth]{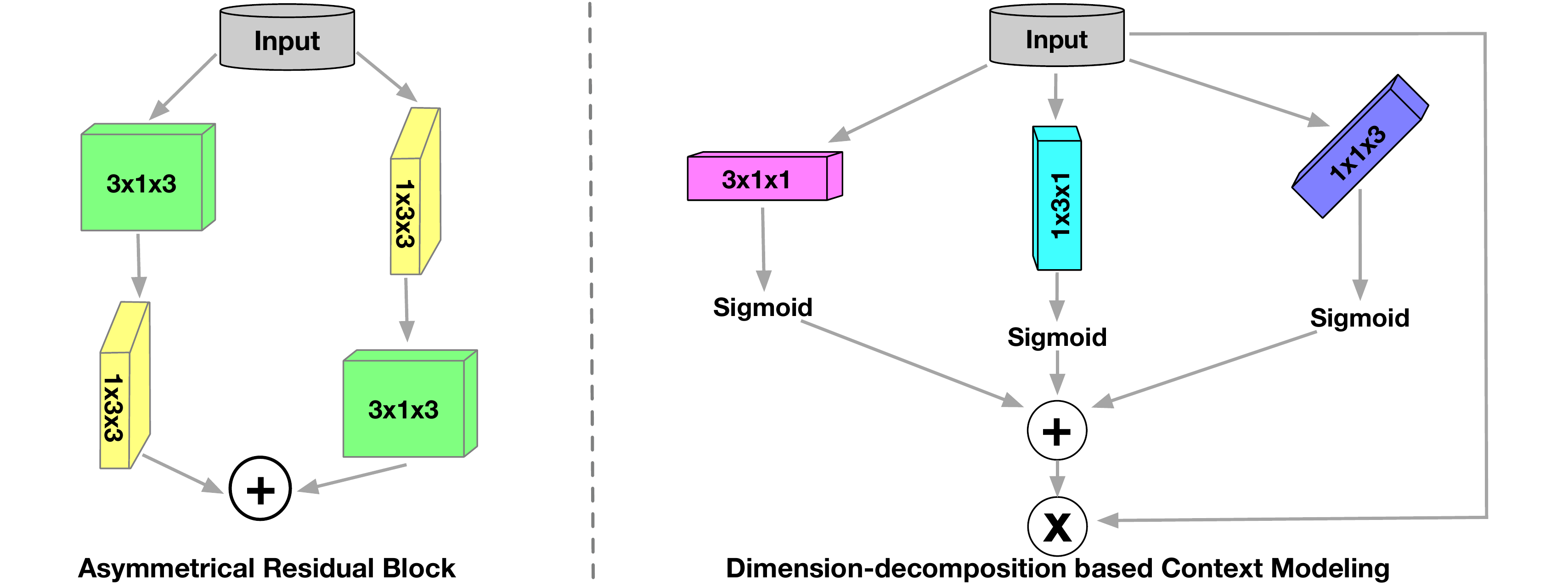}
    \caption{The detailed framework of asymmetry residual block and dimension-decomposition based context modeling.}
    \label{fig:modules}
    \end{figure*}

\subsection{Dimension-decomposition based Context Modeling}

Due to the large varieties of context (for 3D space, its context varies from point cloud to point cloud and should have large diversity), the context tensor should be high-rank~\cite{zhang2019co,wanli2020} to have enough capacity for encoding context information. To model this context feature requires a huge cost, especially in the 3D space, because of the high-rank property of context. Inspired by the high-rank matrix decomposition theory, we can separate the high-rank context into several low-rank representation. In our task, this high-rank context can be divided into three dimensions, \ie, height, width and depth, where all three fragments are both low-rank.
Then we build up the complete high-rank context using these fragments. In this way, this decomposite-aggregate strategy tackles the high-rank difficulty from different views with low-rank constraints. As shown in Fig.~\ref{fig:overall}(bottom), three rank-1 kernels (\ie, $3\times1\times1$, $1\times3\times1$ and $1\times1\times3$) are used to generate these low-rank encoding in all three dimensions. Then the Sigmoid function modulate the convolution results and generates weights for each dimension, in which the co-occurrence contextual information is mined based on the rank-1 tensors from different views. We aggregate all three low-rank activations to obtain the summation to represent the complete context features.

\subsection{Network Optimization}

In this section, we give the details of three parts: 3D cylinder partition, 3D segmentation backbone and segmentation head, as shown in Fig.\ref{fig:overall}. 
Cylinder partition utilizes a 4-layer MLP network with BatchNorm and ReLU to extract point features for each point and select the maximum magnitude of the point features as voxel representation. Our 3D segmentation backbone is derived from U-Net, where 3D convolution is the sparse convolution adapted from~\cite{yan2018second}. As mentioned above, we replace traditional residual block with asymmetrical residual block and insert a DDCM module before the final prediction. The input to segmentation backbone is $C \times H \times W \times L$ tensor. The third part is segmentation head, in which we adapts a 3d convolution layer with $3 \times 3 \times 3$ kernel as a light-weight segmentation head. After the whole pipeline, voxel based prediction, whose size is $ Class \times H \times W \times L$, is obtained. 

For network optimization, we use a weighted cross-entropy loss  and a lovasz-softmax\citep{berman2018lovasz} loss to maximize the point accuracy and the intersection-over-union score for classes. Two losses share the same weight. Thus, the total loss is: $ \zeta_{all} = \zeta_{iou} + \zeta_{acc} $. For the optimizer, Adam with an initial learning rate of 0.001, is employed.

\section{Experiments}


\subsection{Dataset and Metric}

SemanticKITTI \citep{behley2019semantickitti} is a large-scale outdoor-scene dataset for point cloud semantic segmentation. It is derived from the KITTI Vision Odometry Benchmark and collected in Germany with a Velodyne-HDLE64 LiDAR. The dataset consists of 22 sequences, splitting sequences 00 to 10 as training set, and 11 to 21 as test set. Overall, the dataset provides 23201 point clouds for training and 20351 for testing. Following previous literature, sequence 08 is used as the validation set. The dataset has in total 28 classes, where 6 classes are duplicated with moving or non-moving attribute. After merging classes with different moving status and ignore classes with very few points, 19 classes are remained for training and evaluation. To evaluate the proposed method, we leverage mean intersection-over-union (mIoU) metric defined in \citep{behley2019semantickitti} over all classes, given by:
$
IoU_{i} = \frac{TP_{i}}{TP_{i} + FP_{i} + FN_{i}}
$
where $TP_{i}, FP_{i}, FN_{i}$ represent true positive, false positive, and false negative predictions for
class $i$ and the mIoU is the mean value of $IoU_{i}$ over all classes.

\subsection{Backbone and Representation Odyssey from 2D to 3D}
\label{sec:exp_baseline}

For LiDAR segmentation in outdoor scene, there exists many previous literatures, in which various partitions and backbones are proposed among 2D and 3D space. We choose two published cutting-edge networks and some variants with different partitions and backbones (among 2D and 3D space) as a reference group, and conducts extensive experiments to show the odyssey of our network design.

\textbf{Spherical Projection}, RangeNet++~\citep{milioto2019rangenet++} is one of typical methods of spherical projection, which projects point cloud onto a spherical surface surrounding the sensor. Compared with other analogous methods, such as Darknet~\citep{behley2019semantickitti} and SqueezeSeg~\citep{wu2018squeezeseg}, it achieves the best performance on Semantic-Kitti test set. Thus, we choose RangeNet53 as spherical projection baseline and replace original rangenet53 with deeplab-resnet101. For a fairer comparison, we also adopt a KNN as postprocessing method to reduce spatial boundary effect for spherical projection.

\textbf{Polar Bird View Projection}, PolarNet~\citep{zhang2020polarnet} is not traditional bird-view method defined in Cartesian coordinates. It introduces polar coordinates on radius-theta plane to effectively represent these points. Radius-theta encoding can reduce learning complexity due to its small input size. We follow the polar image setting in PolarNet and show the results with different network architectures.

\textbf{Cuboid 3D Voxelization} is a common point representation used in LiDAR segmentation. It converts  point cloud into 3D voxels in Cartesian coordinates. These methods often possess huge computing costs because of the large cuboid voxel resolution and 3D convolution backbone.

\textbf{Cylinder 3D Voxelization} is proposed in this paper. It divides point cloud into small grids in Cylindrical coordinate system. As we claim in section \ref{sec:methods}, cylinder voxel partition meets the varying sparsity of driving-scene LiDAR point cloud and balances point distribution.

\textbf{Analysis.} as shown in Table \ref{table_backbone}, we conduct extensive experiments to evaluate different projections with different segmentation backbones among 2D and 3D space. 
It can be observed that for 2D projections, polar projection outperforms spherical projection methods with different segmentation backbones, such as Resnet-50-FCN, DRN-DeepLab and Resnet101-DeepLab, which demonstrates the superiority of polar projection. 
It is worth noting that our 2D and 3D backbones share the same architecture as shown in Fig.~\ref{fig:overall}. The main difference between 2D and 3D backbones is convolution layer, and we instead use 2D Convolution in 2D backbone.
Based on the same polar projection, our 2D backbone outperforms the polarnet by 2.8\% mIoU, which demonstrates the scalability of the proposed model even in 2D space. 
When we replace the polar projection with our cylinder 3D voxelization, our model has a 1.7\% gain because it retains the 3D topology, which indicates the effectiveness of 3D cylinder partition. 
After converting the 2D backbone to the 3D backbone, the proposed Cylinder3D obtains 4.2\% gain and achieves 64.3\% mIoU on val set. It can be observed that 3D convolution based framework significantly boosts the performance compared to the 2D backbone, which demonstrates the cooperation of 3D Cylinder partition and 3D convolution leads to the point cloud segmentation and verifies our conjecture 3D structure is a crucial aspect in LiDAR segmentation.

\begin{table*}[h]
\caption{Quantitative results of backbone on SemanticKITTI val set.}
\label{table_backbone}
\centering
\begin{adjustbox}{width=\textwidth}
\begin{tabular}{|c|c|c|c|c|c|c|c|c|c|c|c|c|c|c|c|c|c|c|c|c|c|}
\hline
\textbf{Projections} & \textbf{Backbones} & \textbf{mIoU} & \rotatebox{90}{car} &  \rotatebox{90}{bicycle} & \rotatebox{90}{motorcycle} & \rotatebox{90}{truck} & \rotatebox{90}{other-vehicle} & \rotatebox{90}{person} & \rotatebox{90}{bicyclist} & \rotatebox{90}{motorcyclist} & \rotatebox{90}{road} & \rotatebox{90}{parking} & \rotatebox{90}{sidewalk} & \rotatebox{90}{other-ground} &
\rotatebox{90}{building} & \rotatebox{90}{fence} & \rotatebox{90}{vegetation} & \rotatebox{90}{trunk} & \rotatebox{90}{terrain} & \rotatebox{90}{pole} & \rotatebox{90}{traffic} \\
\hline
\hline

\multirow{2}{*}{Spherical Projection}
& RangeNet53\citep{milioto2019rangenet++} & 0.528 & 0.910 & 0.250 & 0.471 & 0.407 & 0.255 & 0.452 & 0.629 & 0.000 & 0.938 & 0.465 & \bf{0.819} & 0.002 & 0.858 & 0.542 & 0.842 & 0.529 & 0.727 & 0.532 & 0.400 \\
\cline{2-22}
& Deeplab-Resnet101 & 0.474 & 0.874 & 0.188 & 0.311 & 0.480 & 0.213 & 0.378 & 0.644 & 0.000 & 0.915 & 0.270 & 0.758 & 0.000 & 0.778 & 0.386 & 0.805 & 0.518 & 0.708 & 0.462 & 0.310  \\
\hline
\hline
\multirow{2}{*}{Polar Projection}
& U-Net \citep{zhang2020polarnet} & 0.556 & 0.928 & 0.292 & 0.353 & 0.608 & 0.272 & 0.545 & 0.676 & 0.000 & 0.936 & \bf{0.473} & 0.797 & \bf{0.073} & 0.899 & 0.497 & 0.861 & 0.628 & 0.718 & 0.607 & 0.406  \\
\cline{2-22}
 & Deeplab-Resnet101 & 0.526 & 0.910 & 0.321 & 0.327 & 0.549 & 0.145 & 0.449 & 0.720 & 0.00 & 0.924 & 0.420 & 0.771 & 0.043 & 0.885 & 0.439 & 0.834 & 0.614 & 0.665 & 0.576 & 0.393  \\
 \cline{2-22}
 & our 2D backbone & 0.584 & 0.936 & 0.334 & 0.412 & 0.811 & 0.391 & 0.540 & 0.748 & 0.000 & 0.933 & 0.450 & 0.784 & 0.027 & 0.897 & 0.532 & 0.869 & 0.648 & 0.728 & 0.625 & 0.436  \\
 \hline
 \hline
 \multirow{1}{*}{Cuboid 3D Voxelization}
& our 3D backbone & 0.609 & 0.946 & 0.448 & 0.563 & 0.756 & 0.379 & 0.672 & \bf{0.906} & \bf{0.005} & 0.916 & 0.432 & 0.762 & 0.024 & \bf{0.910} & 0.586 & 0.860 & 0.661 & 0.695 & 0.609 & 0.446  \\
 \hline
 \hline
 \multirow{2}{*}{Cylinder 3D Voxelization}
& our 2D backbone & 0.601 & 0.941 & 0.420 & 0.580 & 0.748 & 0.438 & 0.557 & 0.758 & 0.00 & 0.935 & 0.468 & 0.793 & 0.009 & 0.902 & 0.516 & 0.878 & 0.680 & \bf{0.757} & 0.592 & 0.463  \\
 \cline{2-22}
 & our 3D backbone & \bf{0.643} & \bf{0.963} & \bf{0.498} & \bf{0.694} & \bf{0.843} & \bf{0.506} & \bf{0.719} & 0.880 & 0.000 & \bf{0.944} & 0.394 & 0.809 & 0.01 & 0.905 & \bf{0.589} & \bf{0.881} & \bf{0.681} & 0.755 & \bf{0.632} & \bf{0.502}  \\
 \hline
\end{tabular}
\end{adjustbox}
\end{table*}

\subsection{Results on SemanticKitti}
In this experiment, we report the results of our model on the SemanticKitti test set from official evaluation server.
As shown in Table \ref{table_test}, our method achieves the state-of-the-art on SemanticKitti test set in comparison with existing methods, including RangeNet++~\citep{milioto2019rangenet++}, PolarNet~\citep{zhang2020polarnet}, SqueezeSegv3~\citep{xu2020squeezesegv3}, RandLA-Net~\citep{hu2020randla}, \etal~ The proposed method outperforms other state-of-the-art methods by  at least 6\% in terms of mIoU.

\begin{table*}[h]
\caption{Quantitative results by our proposed method and state-of-the-art LiDAR Segmentation methods on SemantciKitti test set.}
\label{table_test}
\centering
\begin{adjustbox}{width=\textwidth}
\begin{tabular}{|c|c|c|c|c|c|c|c|c|c|c|c|c|c|c|c|c|c|c|c|c|}
\hline
\textbf{Methods} & \textbf{mIoU} & \rotatebox{90}{car} &  \rotatebox{90}{bicycle} & \rotatebox{90}{motorcycle} & \rotatebox{90}{truck} & \rotatebox{90}{other-vehicle} & \rotatebox{90}{person} & \rotatebox{90}{bicyclist} & \rotatebox{90}{motorcyclist} & \rotatebox{90}{road} & \rotatebox{90}{parking} & \rotatebox{90}{sidewalk} & \rotatebox{90}{other-ground} &
\rotatebox{90}{building} & \rotatebox{90}{fence} & \rotatebox{90}{vegetation} & \rotatebox{90}{trunk} & \rotatebox{90}{terrain} & \rotatebox{90}{pole} & \rotatebox{90}{traffic} \\
\hline
\hline
PointNet\citep{qi2017pointnet} & 0.146 & 0.463 & 0.013 & 0.003 &  0.001 &  0.008 &  0.002 &  0.002 & 0.0 &  0.616 &  0.158 &  0.357 &  0.014 & 0.414 & 12.9 & 0.310 & 0.046 & 0.176 & 0.024 & 0.037  \\
\hline
Splatnet\citep{su2018splatnet} & 0.228 & 0.666 & 0.0 & 0.0 & 0.0 & 0.0 & 0.0 & 0.0 & 0.0 & 0.704 & 0.008 & 0.415 & 0.0 & 0.687 & 0.278 & 0.723 & 0.359 & 0.358 & 0.138 & 0.0\\
\hline
TangentConv\citep{tatarchenko2018tangent} & 0.359 & 0.868 & 0.013 & 0.127 & 0.116 & 0.102 & 0.171 & 0.202 & 0.005 & 0.829 & 0.152 & 0.617 & 0.090 & 0.828 & 0.442 & 0.755 & 0.425 & 0.555 & 0.302 & 0.222\\
\hline
Darknet53\citep{behley2019semantickitti} & 0.499 & 0.864 & 0.245 & 0.327 & 0.255 & 0.226 & 0.362 & 0.336 & 0.047 & 0.918 & 0.648 & 0.746 & \bf{0.279} & 0.841 & 0.55 & 0.783 & 0.501 & 0.640 & 0.389 & 0.522 \\
\hline
RandLA-Net\citep{hu2020randla} & 0.503 & 0.940 & 0.198 & 0.214 & \bf{0.427} & 0.387 & 0.475 & 0.488 & 0.046  & 0.904 & 0.569 & 0.679 & 0.155 & 0.811 & 0.497 & 0.783 & 0.603 & 0.590 & 0.442 & 0.381 \\
\hline
RangeNet++\citep{milioto2019rangenet++} & 0.522 & 0.914 & 0.257 & 0.344 & 0.257 & 0.230 & 0.383 &  0.388 & 0.048 & \bf{0.918} & \bf{0.650} & 0.752 & 0.278 & 0.874 & 0.586 & 0.805 & 0.551 & 0.646 & 0.479 & 0.559 \\
\hline
PolarNet\citep{zhang2020polarnet} & 0.543 & 0.938 & 0.403 & 0.301 & 0.229 & 0.285 & 0.432 & 0.402 & 0.056 & 0.908 & 0.617 & 0.744 & 0.217 & \bf{0.900} & 0.613 & 0.840 & 0.655 & 0.678 & 0.518 & 0.575  \\
\hline
SqueezeSegv3\citep{xu2020squeezesegv3} & 0.559 & 0.925 & 0.387 & 0.365 & 0.296 & 0.330 & 0.456 & 0.462 & \bf{0.201} & 0.917 & 0.634 & 0.748 & 0.264 & 0.89 & 0.594 & 0.82 & 0.587 & 0.654 & 0.496 & 0.589  \\
\hline
Cylinder3D & \bf{0.618} & \bf{0.961} & \bf{0.542} & \bf{0.476} & 0.386 & \bf{0.450} & \bf{0.651} & \bf{0.635} & 0.136 & 0.912 & 0.622 & \bf{0.752} & 0.187 & 0.896 & \bf{0.616} & \bf{0.854} & \bf{0.697} & \bf{0.693} & \bf{0.626} & \bf{0.647}  \\
 \hline
\end{tabular}
\end{adjustbox}
\end{table*}

\subsection{Effects of network components}

In this experiment, we perform the ablation studies to investigate the effects of different network components in Cylinder3D, including Asymmetry residual block, Dimension-decomposition based context modeling and Flip test (a common technique to boost the performance). 
We use the Cylinder partition and 3D U-net (similar to our 3D networks, but use the common residual block and no dimension-decomposition context modeling) as the baseline method. Then we gradually add these network components to observe its effectiveness.
By replacing residual 
block with asymmetry residual block, it can be found about 1.5\% mIoU performance gain is achieved. When adding Dimension-decomposition based Context Modeling, our proposed Cylinder3D achieves 64.3\% in terms of mIoU. 
Moreover, by further incorporating the Flip Test, \ie, flipping the original point cloud via x-axis, y-axis and x-y-axis, and averaging four predictions as the final results, the mIoU increases by another 0.9\%. From the ablation, we can find that both two designed modules achieve the consistent performance gain. 

\begin{table*}[h]
\caption{Effects of network components on SemanticKITTI val set.}
\label{table_net_components}
\centering
\begin{tabular}{c c c c|c}
\hline
\textbf{Baseline} & \textbf{Asymmetry residual block} & \textbf{DDCM} & \textbf{Flip Test} & \textbf{mIoU} \\
\hline
\hline
\checkmark & & & & 0.615 \\
\checkmark & \checkmark & & & 0.630 \\
\checkmark & \checkmark & \checkmark & & 0.643 \\
\checkmark & \checkmark & \checkmark & \checkmark & 0.652 \\
\hline
\end{tabular}
\end{table*}

\subsection{Visualization}

\begin{figure*}[h]
\centering
\begin{tabular}{c}
\includegraphics[width=12cm]{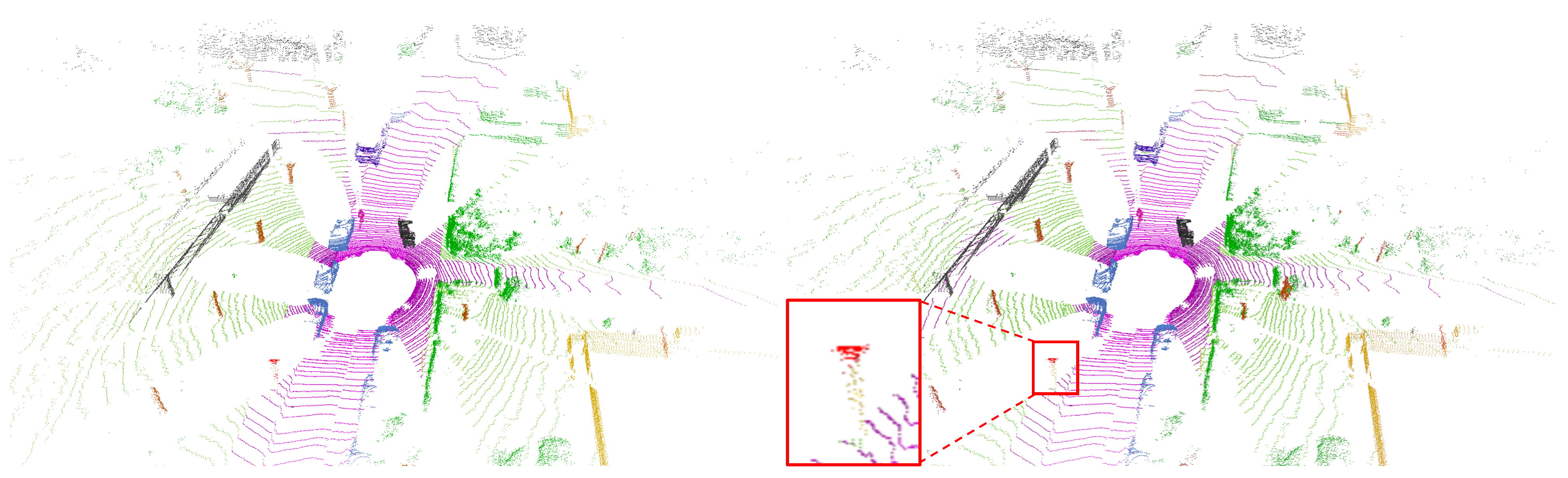}\\
\hline
\includegraphics[width=12cm]{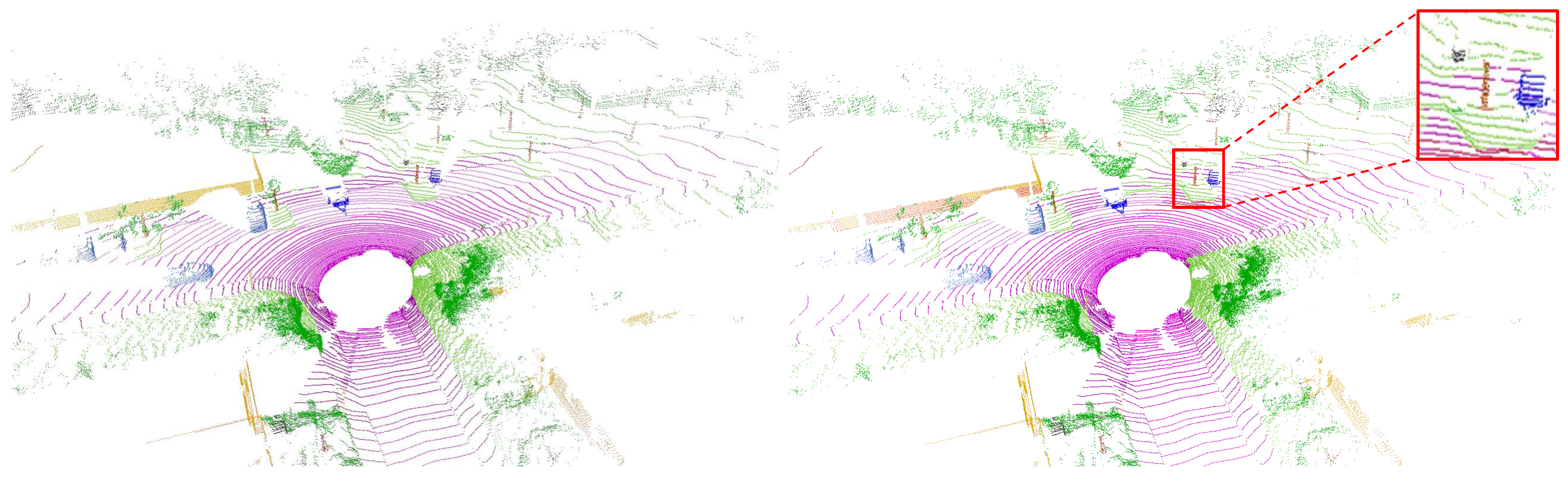} \\
\hline
\includegraphics[width=12cm]{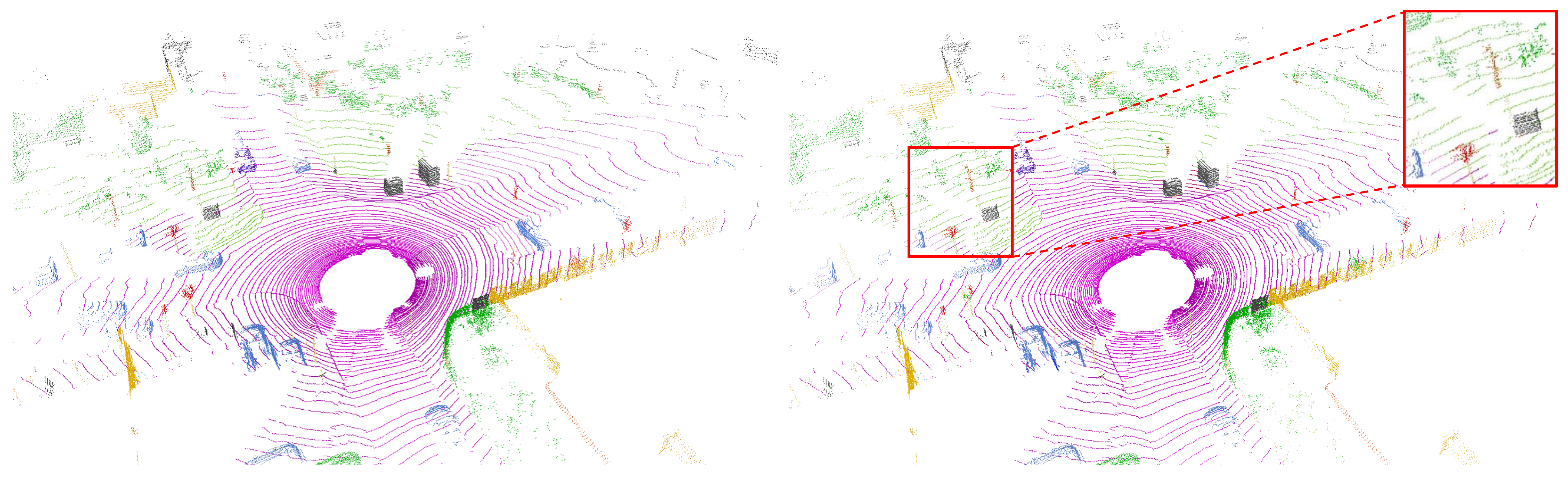} \\
\end{tabular}
\caption{Visualization on validation set. The left is ground-truth and right is our prediction.}
\label{fig:vis_result}
\end{figure*}

Some of the results are visualized in Fig.\ref{fig:vis_result}. It can be observe that the proposed Cylinder3D mainly achieves decent accuracy, and well separates the nearby objects because it maintains the 3D topology and utilizes the geometric information (we highlight corresponding regions with red rectangles).

\section{Conclusion}

In this paper, we follow the 3D nature of lidar point cloud to reposition the focus of lidar segmentation, from 2D to 3D representation and network.
We design a 3D pointcloud representation, named Cylinder partition,
which suits for the varying sparsity of driving-scene lidar point cloud, and propose a 3D convolution based network, where two basic network modules,called Asymmetric Residual Block and Dimension-decomposition based Context Modeling, are introduced to reduce computational cost and explore the high-rank context. With the cooperation of Cylinder partition and 3D convolution networks, our method achieves the state-of-the-art on the SemanticKITTI test set.


\bibliography{example}  

\begin{thebibliography}{34}
\providecommand{\natexlab}[1]{#1}
\providecommand{\url}[1]{\texttt{#1}}
\expandafter\ifx\csname urlstyle\endcsname\relax
  \providecommand{\doi}[1]{doi: #1}\else
  \providecommand{\doi}{doi: \begingroup \urlstyle{rm}\Url}\fi

\bibitem[Wu et~al.(2018)Wu, Wan, Yue, and Keutzer]{wu2018squeezeseg}
B.~Wu, A.~Wan, X.~Yue, and K.~Keutzer.
\newblock Squeezeseg: Convolutional neural nets with recurrent crf for
  real-time road-object segmentation from 3d lidar point cloud.
\newblock In \emph{2018 IEEE International Conference on Robotics and
  Automation (ICRA)}, pages 1887--1893. IEEE, 2018.

\bibitem[Zhang et~al.(2020)Zhang, Zhou, David, Yue, Xi, Gong, and
  Foroosh]{zhang2020polarnet}
Y.~Zhang, Z.~Zhou, P.~David, X.~Yue, Z.~Xi, B.~Gong, and H.~Foroosh.
\newblock Polarnet: An improved grid representation for online lidar point
  clouds semantic segmentation.
\newblock In \emph{Proceedings of the IEEE/CVF Conference on Computer Vision
  and Pattern Recognition}, pages 9601--9610, 2020.

\bibitem[Graham et~al.(2018)Graham, Engelcke, and Van Der~Maaten]{graham20183d}
B.~Graham, M.~Engelcke, and L.~Van Der~Maaten.
\newblock 3d semantic segmentation with submanifold sparse convolutional
  networks.
\newblock In \emph{Proceedings of the IEEE conference on computer vision and
  pattern recognition}, pages 9224--9232, 2018.

\bibitem[Milioto et~al.(2019)Milioto, Vizzo, Behley, and
  Stachniss]{milioto2019rangenet++}
A.~Milioto, I.~Vizzo, J.~Behley, and C.~Stachniss.
\newblock Rangenet++: Fast and accurate lidar semantic segmentation.
\newblock In \emph{2019 IEEE/RSJ International Conference on Intelligent Robots
  and Systems (IROS)}, pages 4213--4220. IEEE, 2019.

\bibitem[{\c{C}}i{\c{c}}ek et~al.(2016){\c{C}}i{\c{c}}ek, Abdulkadir, Lienkamp,
  Brox, and Ronneberger]{cciccek20163d}
{\"O}.~{\c{C}}i{\c{c}}ek, A.~Abdulkadir, S.~S. Lienkamp, T.~Brox, and
  O.~Ronneberger.
\newblock 3d u-net: learning dense volumetric segmentation from sparse
  annotation.
\newblock In \emph{International conference on medical image computing and
  computer-assisted intervention}, pages 424--432. Springer, 2016.

\bibitem[Qi et~al.(2017)Qi, Su, Mo, and Guibas]{qi2017pointnet}
C.~R. Qi, H.~Su, K.~Mo, and L.~J. Guibas.
\newblock Pointnet: Deep learning on point sets for 3d classification and
  segmentation.
\newblock In \emph{Proceedings of the IEEE conference on computer vision and
  pattern recognition}, pages 652--660, 2017.

\bibitem[Thomas et~al.(2019)Thomas, Qi, Deschaud, Marcotegui, Goulette, and
  Guibas]{thomas2019kpconv}
H.~Thomas, C.~R. Qi, J.-E. Deschaud, B.~Marcotegui, F.~Goulette, and L.~J.
  Guibas.
\newblock Kpconv: Flexible and deformable convolution for point clouds.
\newblock In \emph{Proceedings of the IEEE International Conference on Computer
  Vision}, pages 6411--6420, 2019.

\bibitem[Wu et~al.(2019)Wu, Qi, and Fuxin]{wu2019pointconv}
W.~Wu, Z.~Qi, and L.~Fuxin.
\newblock Pointconv: Deep convolutional networks on 3d point clouds.
\newblock In \emph{Proceedings of the IEEE Conference on Computer Vision and
  Pattern Recognition}, pages 9621--9630, 2019.

\bibitem[Wang et~al.(2019)Wang, Sun, Liu, Sarma, Bronstein, and
  Solomon]{wang2019dynamic}
Y.~Wang, Y.~Sun, Z.~Liu, S.~E. Sarma, M.~M. Bronstein, and J.~M. Solomon.
\newblock Dynamic graph cnn for learning on point clouds.
\newblock \emph{Acm Transactions On Graphics (tog)}, 38\penalty0 (5):\penalty0
  1--12, 2019.

\bibitem[Veli{\v{c}}kovi{\'c} et~al.(2017)Veli{\v{c}}kovi{\'c}, Cucurull,
  Casanova, Romero, Lio, and Bengio]{velivckovic2017graph}
P.~Veli{\v{c}}kovi{\'c}, G.~Cucurull, A.~Casanova, A.~Romero, P.~Lio, and
  Y.~Bengio.
\newblock Graph attention networks.
\newblock \emph{arXiv preprint arXiv:1710.10903}, 2017.

\bibitem[Lyu et~al.(2020)Lyu, Huang, and Zhang]{lyu2020learning}
Y.~Lyu, X.~Huang, and Z.~Zhang.
\newblock Learning to segment 3d point clouds in 2d image space.
\newblock In \emph{Proceedings of the IEEE/CVF Conference on Computer Vision
  and Pattern Recognition}, pages 12255--12264, 2020.

\bibitem[Engelmann et~al.(2020)Engelmann, Bokeloh, Fathi, Leibe, and
  Nie{\ss}ner]{engelmann20203d}
F.~Engelmann, M.~Bokeloh, A.~Fathi, B.~Leibe, and M.~Nie{\ss}ner.
\newblock 3d-mpa: Multi-proposal aggregation for 3d semantic instance
  segmentation.
\newblock In \emph{Proceedings of the IEEE/CVF Conference on Computer Vision
  and Pattern Recognition}, pages 9031--9040, 2020.

\bibitem[Zhang et~al.(2020)Zhang, Zhu, Zheng, and Xu]{zhang2020fusion}
J.~Zhang, C.~Zhu, L.~Zheng, and K.~Xu.
\newblock Fusion-aware point convolution for online semantic 3d scene
  segmentation.
\newblock In \emph{Proceedings of the IEEE/CVF Conference on Computer Vision
  and Pattern Recognition}, pages 4534--4543, 2020.

\bibitem[Qi et~al.(2017)Qi, Yi, Su, and Guibas]{qi2017pointnet++}
C.~R. Qi, L.~Yi, H.~Su, and L.~J. Guibas.
\newblock Pointnet++: Deep hierarchical feature learning on point sets in a
  metric space.
\newblock In \emph{Advances in neural information processing systems}, pages
  5099--5108, 2017.

\bibitem[Behley et~al.(2019)Behley, Garbade, Milioto, Quenzel, Behnke,
  Stachniss, and Gall]{behley2019semantickitti}
J.~Behley, M.~Garbade, A.~Milioto, J.~Quenzel, S.~Behnke, C.~Stachniss, and
  J.~Gall.
\newblock Semantickitti: A dataset for semantic scene understanding of lidar
  sequences.
\newblock In \emph{Proceedings of the IEEE International Conference on Computer
  Vision}, pages 9297--9307, 2019.

\bibitem[Wu et~al.(2019)Wu, Zhou, Zhao, Yue, and Keutzer]{wu2019squeezesegv2}
B.~Wu, X.~Zhou, S.~Zhao, X.~Yue, and K.~Keutzer.
\newblock Squeezesegv2: Improved model structure and unsupervised domain
  adaptation for road-object segmentation from a lidar point cloud.
\newblock In \emph{2019 International Conference on Robotics and Automation
  (ICRA)}, pages 4376--4382. IEEE, 2019.

\bibitem[Wang et~al.(2020)Wang, Zhu, and Lin]{wang2020reconfigurable}
T.~Wang, X.~Zhu, and D.~Lin.
\newblock Reconfigurable voxels: A new representation for lidar-based point
  clouds.
\newblock \emph{arXiv preprint arXiv:2004.02724}, 2020.

\bibitem[Han et~al.(2020)Han, Zheng, Xu, and Fang]{han2020occuseg}
L.~Han, T.~Zheng, L.~Xu, and L.~Fang.
\newblock Occuseg: Occupancy-aware 3d instance segmentation.
\newblock In \emph{Proceedings of the IEEE/CVF Conference on Computer Vision
  and Pattern Recognition}, pages 2940--2949, 2020.

\bibitem[Zhu et~al.(2020)Zhu, Ma, Wang, Xu, Shi, and Lin]{zhu2020ssn}
X.~Zhu, Y.~Ma, T.~Wang, Y.~Xu, J.~Shi, and D.~Lin.
\newblock Ssn: Shape signature networks for multi-class object detection from
  point clouds.
\newblock \emph{arXiv preprint arXiv:2004.02774}, 2020.

\bibitem[Tchapmi et~al.(2017)Tchapmi, Choy, Armeni, Gwak, and
  Savarese]{tchapmi2017segcloud}
L.~Tchapmi, C.~Choy, I.~Armeni, J.~Gwak, and S.~Savarese.
\newblock Segcloud: Semantic segmentation of 3d point clouds.
\newblock In \emph{2017 international conference on 3D vision (3DV)}, pages
  537--547. IEEE, 2017.

\bibitem[Long et~al.(2015)Long, Shelhamer, and Darrell]{long2015fully}
J.~Long, E.~Shelhamer, and T.~Darrell.
\newblock Fully convolutional networks for semantic segmentation.
\newblock In \emph{Proceedings of the IEEE conference on computer vision and
  pattern recognition}, pages 3431--3440, 2015.

\bibitem[Ronneberger et~al.(2015)Ronneberger, Fischer, and
  Brox]{ronneberger2015u}
O.~Ronneberger, P.~Fischer, and T.~Brox.
\newblock U-net: Convolutional networks for biomedical image segmentation.
\newblock In \emph{International Conference on Medical image computing and
  computer-assisted intervention}, pages 234--241. Springer, 2015.

\bibitem[Chen et~al.(2017)Chen, Papandreou, Kokkinos, Murphy, and
  Yuille]{chen2017deeplab}
L.-C. Chen, G.~Papandreou, I.~Kokkinos, K.~Murphy, and A.~L. Yuille.
\newblock Deeplab: Semantic image segmentation with deep convolutional nets,
  atrous convolution, and fully connected crfs.
\newblock \emph{IEEE transactions on pattern analysis and machine
  intelligence}, 40\penalty0 (4):\penalty0 834--848, 2017.

\bibitem[Chen et~al.(2018)Chen, Zhu, Papandreou, Schroff, and
  Adam]{chen2018encoder}
L.-C. Chen, Y.~Zhu, G.~Papandreou, F.~Schroff, and H.~Adam.
\newblock Encoder-decoder with atrous separable convolution for semantic image
  segmentation.
\newblock In \emph{Proceedings of the European conference on computer vision
  (ECCV)}, pages 801--818, 2018.

\bibitem[Zhao et~al.(2017)Zhao, Shi, Qi, Wang, and Jia]{zhao2017pyramid}
H.~Zhao, J.~Shi, X.~Qi, X.~Wang, and J.~Jia.
\newblock Pyramid scene parsing network.
\newblock In \emph{Proceedings of the IEEE conference on computer vision and
  pattern recognition}, pages 2881--2890, 2017.

\bibitem[Wang et~al.(2019)Wang, Xie, Li, Hou, Lu, Yu, and Shao]{wang2019shape}
W.~Wang, E.~Xie, X.~Li, W.~Hou, T.~Lu, G.~Yu, and S.~Shao.
\newblock Shape robust text detection with progressive scale expansion network.
\newblock In \emph{Proceedings of the IEEE Conference on Computer Vision and
  Pattern Recognition}, pages 9336--9345, 2019.

\bibitem[Zhang et~al.(2019)Zhang, Zhang, Wang, and Xie]{zhang2019co}
H.~Zhang, H.~Zhang, C.~Wang, and J.~Xie.
\newblock Co-occurrent features in semantic segmentation.
\newblock In \emph{Proceedings of the IEEE Conference on Computer Vision and
  Pattern Recognition}, pages 548--557, 2019.

\bibitem[Chen et~al.(2020)Chen, Zhu, Sun, He, Li, Shen, and Yu]{wanli2020}
W.~Chen, X.~Zhu, R.~Sun, J.~He, R.~Li, X.~Shen, and B.~Yu.
\newblock Tensor low-rank reconstruction for semantic segmentation.
\newblock \emph{arXiv preprint arXiv:2008.00490}, 2020.

\bibitem[Yan et~al.(2018)Yan, Mao, and Li]{yan2018second}
Y.~Yan, Y.~Mao, and B.~Li.
\newblock Second: Sparsely embedded convolutional detection.
\newblock \emph{Sensors}, 18\penalty0 (10):\penalty0 3337, 2018.

\bibitem[Berman et~al.(2018)Berman, Rannen~Triki, and
  Blaschko]{berman2018lovasz}
M.~Berman, A.~Rannen~Triki, and M.~B. Blaschko.
\newblock The lov{\'a}sz-softmax loss: A tractable surrogate for the
  optimization of the intersection-over-union measure in neural networks.
\newblock In \emph{Proceedings of the IEEE Conference on Computer Vision and
  Pattern Recognition}, pages 4413--4421, 2018.

\bibitem[Xu et~al.(2020)Xu, Wu, Wang, Zhan, Vajda, Keutzer, and
  Tomizuka]{xu2020squeezesegv3}
C.~Xu, B.~Wu, Z.~Wang, W.~Zhan, P.~Vajda, K.~Keutzer, and M.~Tomizuka.
\newblock Squeezesegv3: Spatially-adaptive convolution for efficient
  point-cloud segmentation.
\newblock \emph{arXiv preprint arXiv:2004.01803}, 2020.

\bibitem[Hu et~al.(2020)Hu, Yang, Xie, Rosa, Guo, Wang, Trigoni, and
  Markham]{hu2020randla}
Q.~Hu, B.~Yang, L.~Xie, S.~Rosa, Y.~Guo, Z.~Wang, N.~Trigoni, and A.~Markham.
\newblock Randla-net: Efficient semantic segmentation of large-scale point
  clouds.
\newblock In \emph{Proceedings of the IEEE/CVF Conference on Computer Vision
  and Pattern Recognition}, pages 11108--11117, 2020.

\bibitem[Su et~al.(2018)Su, Jampani, Sun, Maji, Kalogerakis, Yang, and
  Kautz]{su2018splatnet}
H.~Su, V.~Jampani, D.~Sun, S.~Maji, E.~Kalogerakis, M.-H. Yang, and J.~Kautz.
\newblock Splatnet: Sparse lattice networks for point cloud processing.
\newblock In \emph{Proceedings of the IEEE Conference on Computer Vision and
  Pattern Recognition}, pages 2530--2539, 2018.

\bibitem[Tatarchenko et~al.(2018)Tatarchenko, Park, Koltun, and
  Zhou]{tatarchenko2018tangent}
M.~Tatarchenko, J.~Park, V.~Koltun, and Q.-Y. Zhou.
\newblock Tangent convolutions for dense prediction in 3d.
\newblock In \emph{Proceedings of the IEEE Conference on Computer Vision and
  Pattern Recognition}, pages 3887--3896, 2018.

\end{thebibliography}

\end{document}